\documentclass[letterpaper, 10 pt, conference]{ieeeconf}  

\IEEEoverridecommandlockouts                              

\usepackage{hyperref} 
\usepackage{mathrsfs}
\usepackage{amsfonts}
\usepackage{amsmath}
\usepackage[flushleft]{threeparttable}
\usepackage{tablefootnote}
\usepackage{multirow}
\usepackage{graphicx}
\usepackage{subfigure}
\usepackage{hanging}
\usepackage{color}
\usepackage{tikz}
\usetikzlibrary{calc,arrows,decorations.markings}

\graphicspath{{../},{../image}}

\newcommand{\PSMC}{PS-CNN$_{\textrm{MC}}$}
\newcommand{\PSTS}{PS-CNN$_{\textrm{TS}}$}

\title{\LARGE \bf
Using Synthetic Data and Deep Networks \\ to Recognize Primitive Shapes for
Object Grasping}

\author{Yunzhi Lin$^{1}$$^{\ast}$, Chao Tang$^{1}$$^{\ast}$, Fu-Jen Chu$^{1}$, 
        and Patricio A.  Vela $^{1}$
\thanks{$^{1}$ Y. Lin, C. Tang, F. Chu, and P.A. Vela are with the
School of Electrical and Computer Engineering, and the
Institute for Robotics and Intelligent Machines, Georgia Institute of
Technology, Atlanta, GA.  
{\tt\small \{ctang84,ylin466,fujenchu,pvela\}@gatech.edu}}%
\thanks{This work was supported in part by NSF Award \#1605228}%
\thanks{$\ast$ The first two authors contributed equally.}%
}

\begin{document}

\maketitle
\thispagestyle{empty}
\pagestyle{empty}


%
%
\begin{abstract}
A segmentation-based architecture is proposed to decompose objects into
multiple primitive shapes from monocular depth input for robotic
manipulation. The backbone deep network is trained on synthetic data with 6
classes of primitive shapes generated by a simulation engine. 
Each primitive shape is designed with parametrized grasp families,
permitting the pipeline to identify multiple grasp candidates per shape
primitive region. The grasps are priority ordered via proposed ranking algorithm, with the first feasible
one chosen for execution. On task-free grasping of individual objects, the
method achieves a 94\% success rate. On task-oriented grasping, it achieves
a 76\% success rate. 
Overall, the method supports the hypothesis that shape primitives can support
task-free and task-relevant grasp prediction. 
\end{abstract}

\section{Introduction}
%
Manipulation is a multi-step task consisting of sequential actions applied
to an object, including: perception, path planning, and
closure of the gripper, followed by a task-relevant motion with the grasped
object \cite{Ciocarlie_Hsiao_Jones_Chitta_Rusu_Şucan_2014}. 
Due to the diversity of objects that a robotic arm could grasp, the
grasping process remains an open problem in the field of robotics.
Analytical or model-based approaches
have trouble addressing this diversity.  Recent research has turned to deep
learning as a robust means to score or detect grasps.

%
Deep learning is a data-driven approach typically requiring large datasets
to recover the desired input/output function, which for grasping is often
an image/grasp pair (permitting multiple grasps per image). 
Manual annotation for dataset creation tends to be limited in volume
\cite{cornellgrasp}, leading roboticists to exploit 
robotic simulators \cite{mahler2017dex,depierre2018jacquard} 
or actual deployment \cite{levine2018learning}  
for automatic generation of training data.
Many of these emphasize the grasp as opposed to the object, with good
reason as object-centric approaches would require creating 3D models or
scanning real objects in large volume \cite{wohlkinger20123dnet,calli2017yale}, 
and concomitantly require a high accuracy detector. 
%
Instead, the deep networks aim to generalize across object classes to
establish general grasping rules from visual evidence.  Implicitly, the
deep network is to learn the notion of shape within its internal feature
representation.


%
This paper proposes to complement that idea with a more geometric and
explicitly shape-centric approach by employing the notion of primitive
shapes.
Primitive shapes offer a powerful means to alleviate the data inefficiency
problem
\cite{yamanobe2010grasp,jain2016grasp,tobin2017domain}
by abstracting target objects to primitive shapes with {\em a priori} known
grasp configurations. Most primitive shape methods represent objects as
a single shape from a small library \cite{jain2016grasp} or apply
model-based rules to deconstruct objects \cite{yamanobe2010grasp}.  As a
result, they do not handle novel shapes having unmodeled geometry or
being the union of primitive shapes.

%
%
Building from Yamanobe \textit{et al.}'s model decomposition idea
\cite{yamanobe2010grasp},
this paper aims to permit a
primitive shapes detector to generalize its grasp strategies to household
objects. 
The detector exploits the state-of-the-art instance segmentation 
deep network Mask R-CNN \cite{he2017mask} trained to segment a depth image
according to the primitive shapes it contains. 
Model-based matching of the primitive shape's point cloud, from a primitive
shape database, recovers candidate grasp families associated to the shape. 
Grasp candidates are ranked then tested in order to find a feasible grasp.


%

%
Synthetic ground truth data based on parametrized sets of primitive shape
classes and their grasp families avoids extensive manual annotation.
The shape classes are sufficiently representative of object parts
associated to household objects yet low enough in cardinality that grasp
family modeling is quick. 
Domain randomization \cite{tobin2017domain,tobin2018domain} 
over the parametrized shapes generates a rich set of ground truth
input/ouput data based on a robotics simulation engine
\cite{rohmer2013v}.  The result is a large synthetic dataset composed
of different primitive shapes combinations, quantities, and layouts. 

%
Modern robotics simulation engines \cite{rohmer2013v,blender} 
render and simulate virtual environments quite cleanly, often leading to
sensor imagery with less defects than real sensors.
Depth sensor like the Kinect v1 loses details and introduce noise during the
depth capture \cite{planche2017depthsynth,Sweeney2019ASA}, leading to a
distribution shift or gap between the simulated data and depth sensor
output. The gap is addressed in a bi-directional manner by lightly
corrupting the simulated data and denoising the depth image data. The
intent is to introduce real sensor artifacts that cannot be removed in the
simulated images, and to denoise the real images to match the simulated
images.



%
Employing the deep network for grasping enables a robotic arm with depth
camera to identify primitive shapes within the scene, and to identify
suitable grasp options for an object from the shapes.
To arrive at the overall system, from training to deployment, the paper
covers the following contributions:

\begin{hangparas}{1em}{1}
- An automated ground truth generation strategy to rapidly generate
  input/output data. It uses 6 classes of primitive shapes with
  parametrized grasp families in concert with a robotics simulation engine
  and domain randomization; 

- A deep network, segmentation-based pipeline first decomposing objects
  into multiple primitive shapes from a depth image, followed by surface
  model-fitting, grasp prioritization/selection via proposed ranking algorithm.

  
  - Experimental testing and evaluation of grasping accuracy using a 7dof
    robotic arm + depth sensor setup, including both standard and 
    task-oriented benchmarking.
\end{hangparas}

\section{Related Work}

%
%
%
%
Grasping is a mechanical process which can be described mathematically given
prior knowledge of the target object's properties (geometry, hardness,
etc.), the hand contact model, and the hand dynamics.
Mechanics-based approaches with analytical solutions work well for some
objects but cannot successfully apply to other, often novel, targets
\cite{Tung_Kak_1996,Prattichizzo_Malvezzi_Gabiccini_Bicchi_2012,Rosales_Suarez_Gabiccini_Bicchi_2012}. 
With advances in machine learning, these methods gave way to purely
data-driven approaches \cite{Bohg_Morales_Asfour_Kragic_2014} or combined
approaches employing analytic scoring with {\em image-to-grasp} learning
\cite{mahler2017dex}.  
Contemporary solutions employ deep learning \cite{caldera2018review} and
leverage available training data. 

%
%
Deep learning strategies primarily take one of three types.
%
%
The first type exploits the strong detection or classification capabilities of
deep networks to recognize candidate structured grasping representations
\cite{Watson_Hughes_Iida_2017,Park_Chun_2018,Chu_Xu_Vela_2018,%
satish2019onpolicy}. The most
common representation is the $SE(2) \times \mathbb{R}^2$ grasp
representation associated to a parallel plate gripper.
As a computer vision problem, recognition accuracy is high (up to around 95\%), 
with a performance drop during robotic implementation (to around 90\% or
less). Training involves image/grasp datasets obtained from
manual annotation \cite{lenz2015deep} or 
simulated grasping \cite{depierre2018jacquard,satish2019onpolicy}. Within
this category there is also a mixed approach, DexNet \cite{mahler2017dex}, using
random sampling and analytical scoring followed by deep network regression
to output refined, learnt grasp quality scores for grasp selection. By
using simulation with an imitation learning methodology, tens of thousands
to millions of annotations support DexNet regression training. 
Success rates vary from 80\% to 93\% depending on the task.
When sufficient resources are available, the second strategy type replaces
simulation with actual experiential data coupled to deep network
reinforcement learning methods \cite{levine2018learning,zeng2018learning}.
Often, methods based on simulation or experience are
configuration-dependent; they learn for specific robot and camera setups.
The third strategy is based on object detection or recognition
\cite{Bohg_Morales_Asfour_Kragic_2014}. Recent work employed deep learning
to detect objects and relative poses to inform grasp planning
\cite{TrEtAl_2018_DeepObjPoseEst}, while another learnt to perform
object-agnostic scene segmentation to differentiate objects
\cite{danielczuk2019segmenting} and aided DexNet grasp selection process.
Like \cite{danielczuk2019segmenting}, this paper focuses on where to find
candidate grasps as opposed to quality scoring candidate grasps.

%
%

\begin{figure}[t]
  \centering
  \vspace*{0.06in}
  \begin{tikzpicture}[inner sep = 0pt, outer sep = 0pt]
    \node[anchor=south west] (fnC) at (0in,0in)
      {\includegraphics[height=0.95in,clip=true,trim=0in 0.25in 0in 0.35in]{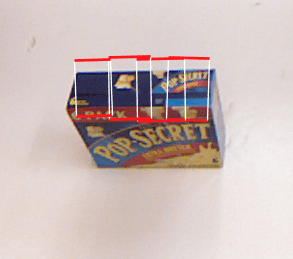}};
    \node[anchor=south west,xshift=2pt] (fnJ) at (fnC.south east)
      {\includegraphics[height=0.95in,clip=true,trim=0in 0.33in 0in 0.27in]{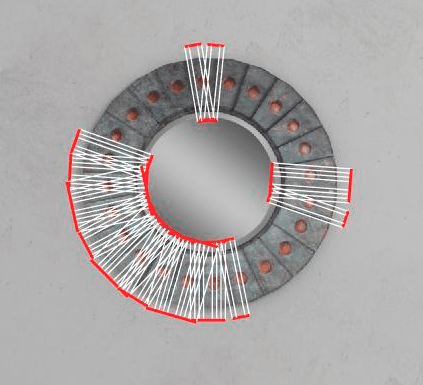}};
    \node[anchor=north west,xshift=2pt,yshift=1em] at (fnC.south west) 
      {\small Cornell};
    \node[anchor=north west,xshift=2pt,yshift=1em] at (fnJ.south west) 
      {\small Jacquard};
  \end{tikzpicture}
  \caption{Grasp annotation data with missing grasp candidates. \label{annotated}}
  \vspace*{-0.25in}
\end{figure}

%
%
%
%
Deep learning grasp methods suffer from two related problems, 
sparse grasp annotations or insufficiently rich data (i.e., covariate shift).  
The former can be seen in Figure \ref{annotated}, which shows an image from
the Cornell dataset \cite{cornellgrasp} and another from 
the Jacquard dataset \cite{depierre2018jacquard}. Both lack annotations in
graspable regions due to missing manual annotation or a false negative in
the simulated scenario (either due to poor sampling or incorrect physics).
Sampling insufficiency can be seen in \cite{satish2019onpolicy},
where the DexNet training policy was augmented with an improved (on-policy)
oracle providing a richer sampling space. Yet, sampling from a continuous
space is bound to under-represent the space of possible options, especially
as the dimension of the parametric grasp space increases. 
This paper proposes to more fully consider shape primitives
\cite{miller2003automatic} due to their known, parametrized grasp families
\cite{yamanobe2010grasp,shiraki2014modeling}.  The parameterized families
provide a continuum of grasp options rather than a sparse sampling. A
complex object can be decomposed into parts representing distinct surface
categories based on established primitives.

%
Deep network approaches for shape primitive segmentation to inform grasping do
not appear to be studied. Past research explored shape primitive approaches
in the context of traditional point cloud processing and fitting for the
cases of 
superquadric \cite{Goldfeder_Allen_Lackner_Pelossof_2007}
and box surfaces \cite{huebner2008selection}.
Approaches were also proposed to simply model each object as a single primitive
shape \cite{jain2016grasp,fang2018learning}, which did not exploit the
potential of primitive shapes to generalize to unseen/novel objects. 

%
%
%
%
%

%

\newcommand{\depthIm}{\mathcal{D}}
\newcommand{\grasp}{\mathcal{G}}
\newcommand{\primSeg}{\mathcal{P}}
\newcommand{\primSegi}{{\mathcal{P}}_i}
\newcommand{\primSet}{I}
\newcommand{\primInd}{i}
\newcommand{\primShape}{P}
\newcommand{\graspInd}{\alpha}

\section{Grasping from Primitive Shapes Recognition}

\begin{figure*}[ht!]
  \centering
  \begin{tikzpicture}[inner sep = 0pt, outer sep = 0pt]
    \node[anchor=south west] at (0in,0in)
      {{\includegraphics[width=1\textwidth]{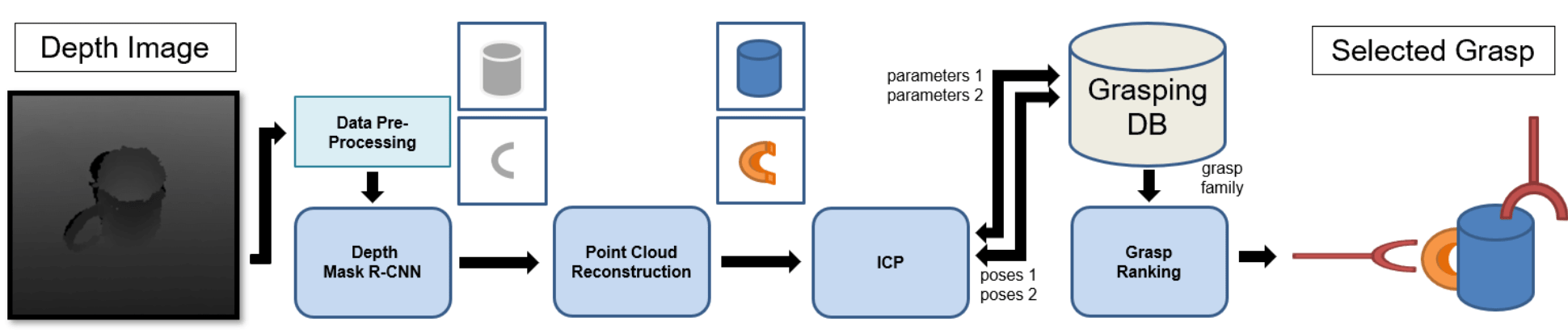}}};
    \node[yshift=-2pt] at (1.67in,0in) {\small (a)};
    \node[yshift=-2pt] at (4in,0in) {\small (b)};
    \node[yshift=-2pt] at (5.17in,0in) {\small (c)};
    \node at (2.6in,1.2in) {$\primSeg_{i_1}$};
    \node at (2.6in,0.8in) {$\primSeg_{i_2}$};
    \node at (3.75in,1.2in) {$\primShape_{j_1}$};
    \node at (3.75in,0.8in) {$\primShape_{j_2}$};
  \end{tikzpicture}
  \vspace*{-.25in}
  \caption{The proposed deep network, segmentation-based pipeline. 
    From monocular depth input, objects are segmented into primitive shape 
    classes, with the object to grasp extracted and converted into
    primitive shape point clouds.  Surface model-fitting, grasp scoring,
    and grasp selection processes follow.
   \textbf{(a)} Depth input is segmented into primitive shapes;
   \textbf{(b)} The best matching shape and pose per primitive shape is
     identified; 
   \textbf{(c)} Candidate grasps are priority ranked and tested for
     feasibility with the first feasible grasp chosen for physical robot
     executions.
   \label{fig:pipeline}}
  \vspace*{0.1in}
  \begin{tikzpicture}[inner sep=0pt, outer sep=0pt]
     \node[anchor=south west] (Full) at (0in,0in)
       {\includegraphics[height=1.10in,clip=true,trim=0.25in 0.25in 0in 0.25in]{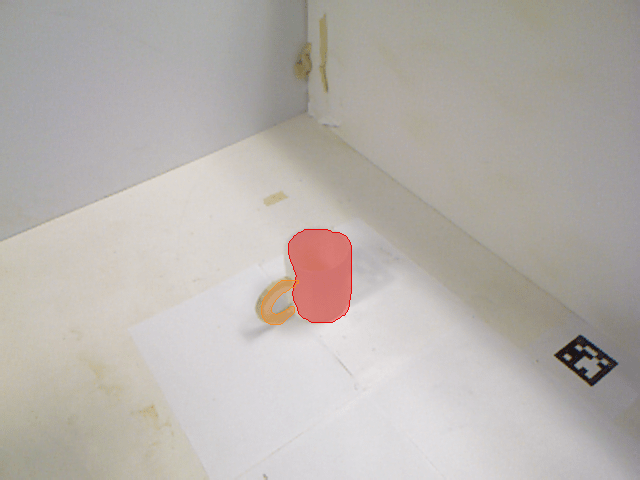}};
     \node[anchor=south west,xshift=3pt,yshift=1pt] at (Full.south west)
     {\small Setup (no zoom)};

     \node[anchor=north west,xshift=2pt] (SS) at (Full.north east)
       {\includegraphics[height=0.535in,clip=true,trim=3.75in 1.2in 2.63in 3.5in]{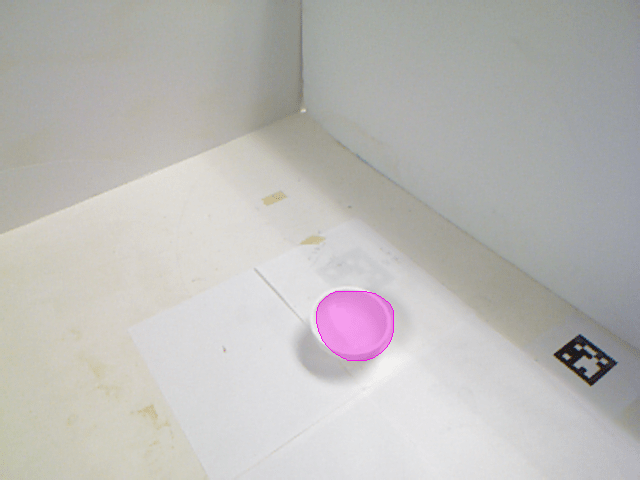}};
     \node[anchor=south west,xshift=2pt] (Cyl) at (Full.south east)
       {\includegraphics[height=0.535in,clip=true,trim=3.2in 1.2in 2.93in 3.3in]{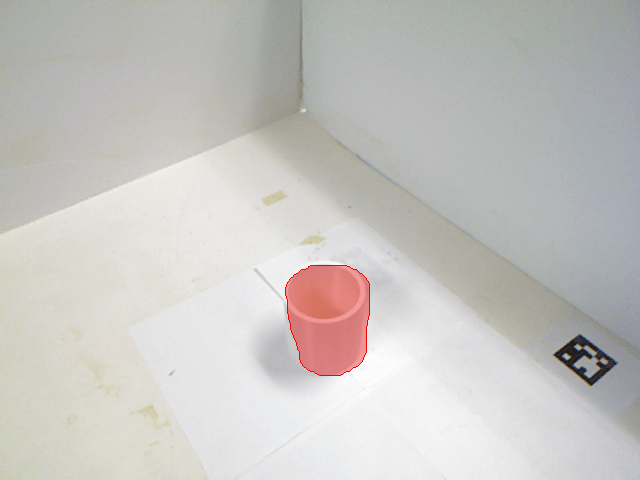}};
      \node[anchor=south west,xshift=2pt] (Bask) at (Cyl.south east)
       {\includegraphics[height=0.535in,clip=true,trim=2.7in 1.6in 3.63in 2.9in]{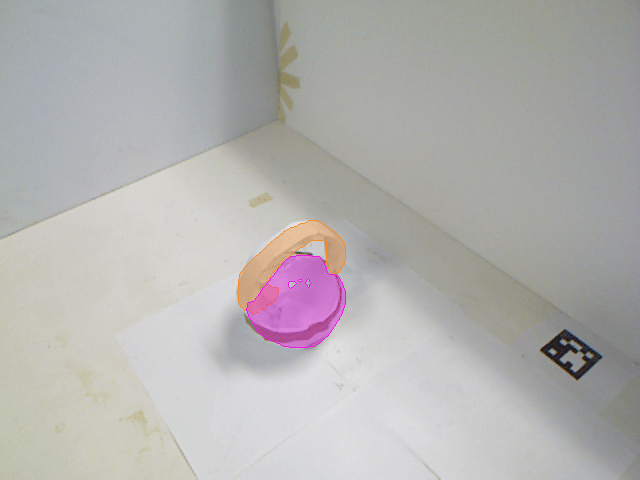}};
%
%
%
%
     \node[anchor=north west,xshift=2pt] (Pot) at (SS.north east)
       {\includegraphics[height=0.535in,clip=true,trim=3.5in 1.5in 2.83in 3.0in]{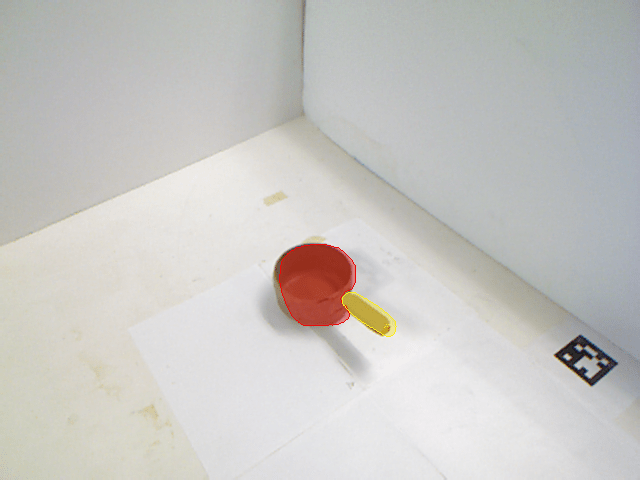}};

     \node[anchor=south east] (M2) at (0.99\textwidth,0in)
       {\includegraphics[height=1.1in,clip=true,trim=1.2in 0in 2.55in 2.9in]{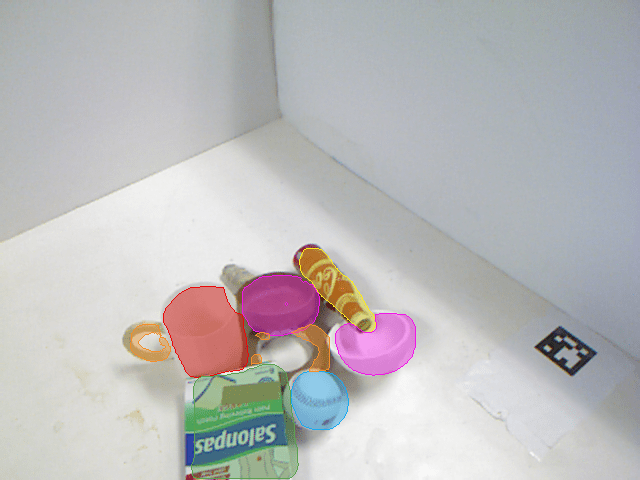}};
     \node[anchor=south east,xshift=-2pt] (M1) at (M2.south west)
       {\includegraphics[height=1.1in,clip=true,trim=2.4in 1.0in 2.4in 1.9in]{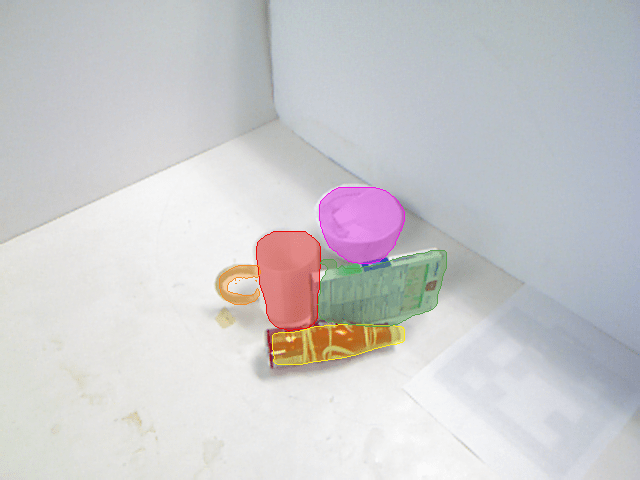}};
     \node[anchor=south east,xshift=-2pt] (M0) at (M1.south west)
       {\includegraphics[height=1.1in,clip=true,trim=2.2in 0in 2.45in 2.9in]{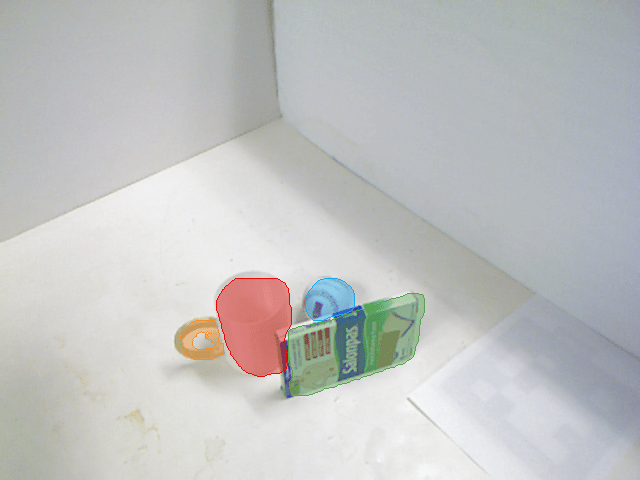}};
   \end{tikzpicture}
   \caption{Sample segmentation outcomes for test scenarios consisting of
   individual and multiple objects (zoomed and cropped images). 
     \label{fig:segdemo}}
  \vspace*{-0.20in}
\end{figure*}

The intent behind this investigation is to explore the potential value of
using deep networks to segment a scene according to the surface primitives
contained within it, in order to establish {\em how} one may grasp.  
Once the object or region to grasp is known, post-processing recovers the
shape geometry and the grasp family associated to the shape. 
The state-of-the-art instance segmentation deep network Mask R-CNN
\cite{he2017mask} serves as the backbone network for converting depth
images into primitive shape segmentation images. 
Importantly, a synthetically generated training set using only shape
primitives in concert with domain randomization \cite{tobin2017domain}
covers a large set of scene visualizations. 
The ability to decompose unseen/novel objects into distinct shape regions,
often with explicitly distinct manipulation affordances, permits
task-oriented grasping \cite{fang2018learning}.

The vision-based robotic grasping problem here presumes the existence of a
depth image $\depthIm$ $\in$ $\mathbb{R}^{H\times W}$ 
($H$ and $W$ are image height and width) capturing a scene containing an
object to grab.  The objective is to abstract the scene into a set of
primitive shapes and generate grasp configurations from them.  
A complete solution involves establishing a routine or process, $f$,
mapping the depth image $\depthIm$ to a grasp 
$\mathcal{G} = f(\mathcal{D}) \in SE(3)$.
The grasp configuration $\grasp \in SE(3)$ specifies the final pose in the
world frame of the end-effector. 

%
Per Figure \ref{fig:pipeline}, the process is divided into three stages. In
the first stage, the depth image $\depthIm$ gets segmented according to
defined primitive shape categories indexed by the set $\primSet$. 
The primitive shape segmentation images are $\primSegi$ for $\primInd \in
\primSet$.
The segmentation $\primSegi$ and the depth image $\depthIm$ generate
segmented point clouds in 3D space for the primitive surfaces attached to
the label $\primInd$. In the second stage, when the grasp target is
established, the surface primitives attached to the target grasp region
are converted into a corresponding set of primitive shapes $\primShape_{j}$
in 3D space, where $j$ indexes the different surface primitive segments. 
In the third stage, the parametrized grasp families of the surface primitives
are used to generate grasp configurations $\grasp_{\graspInd}$ for
$\alpha \in \mathbb{N}^+$.  A prioritization process leads to rank ordered
grasps with the first feasible grasp being the one to execute. This section
details the three stages and the deep network training method.


%
%
\subsection{Primitive Shape Segmentation Using Mask R-CNN}
%
%
The proposed approach hypothesizes that commonly seen household
objects can be decomposed into one or more primitive shapes and represented
by pre-defined shape parameters. After studying several household object
datasets
\cite{calli2017yale,chen2003visual,funkhouser2003search,shilane2004princeton}, 
the set of primitive shapes was decided to be:
{\em cylinder}, 
{\em cuboid}, 
{\em ring}, 
{\em sphere},
{\em semi-sphere}, and 
{\em stick}.

\begin{table}[t]
  \centering
  \vspace*{0.075in}
  \caption{Primitive Shape Classes\label{tab:primtive_shape_design}}
  \vspace*{-0.05in}
  \setlength\tabcolsep{2pt}
  \begin{tabular}{|c|c|c|}
    \hline
    Class  & Parameters  & Shape Template (unit: cm) \\ 
    \hline 
    %
    \hline
    \multirow{2}{*}{\parbox{0.495in}{\centering Cylinder \\ (wide, tall)}} & 
    \multirow{2}{*}{$(r_{in},r_{out},h)$}  & 
    \parbox{1.25in}{
      \vspace*{-1.1ex}
      \begin{equation} \nonumber
      \begin{split}
        \sigma & \in [2.2,3.0] \\ 
        \rho_{wide} & = (1, 1.15, 2.45)
      \end{split}
      \end{equation}
      \vspace*{-2.2ex}} 
    \\
    \cline{3-3} 
    & & 
    \parbox{1.25in}{
      \vspace*{-1.1ex}
      \begin{equation} \nonumber
      \begin{split}
        \sigma & \in [2, 6.5] \\ 
        \rho_{tall} & = (1, 1.15, 5.063)
      \end{split}
      \end{equation}
      \vspace*{-2.2ex}} 
    \\ 
    \hline
    Ring & 
      $(r_{in},r_{out},h)$ & 
    \parbox{1.25in}{
      \vspace*{-1.1ex}
      \begin{equation} \nonumber
      \begin{split}
        \sigma & \in [1.85,5.0] \\ 
        \rho   & = (1, 1.25, 0.6125)
      \end{split}
      \end{equation}
    \vspace*{-2.2ex}} 
    \\
    \hline
    {\parbox{0.495in}{\centering Cuboid \\ (wide, tall)}} & 
    $(h,w,d)$ & 
    \parbox{1.30in}{
      \vspace*{-1.1ex}
      \begin{equation} \nonumber
      \begin{split}
        \sigma & \in [2.2,3.0] \\ 
        \rho_{wide} & = (6.25,4,11) \\
        \rho_{tall} & = (18.49,1.9,12.09) 
      \end{split}
      \end{equation} 
    \vspace*{-1.6ex}} \\
    \hline
    Stick & 
    $(r,l)$ & 
      $\sigma \in [5.5,19],\ \rho = (1.25, 1)$\rule{0pt}{2.25ex} 
    \\ \hline
    Semi-sphere & 
    $(r_{in}, r_{out})$ & 
       $\sigma \in [3.5,8],\ \rho = (0.9, 1)$\rule{0pt}{2.25ex}
    \\ \hline
    Sphere & 
      $r$  & $\sigma \in [2.25, 4.5],\ r = 1$ \rule{0pt}{2.25ex}\\ 
    \hline
  \end{tabular}
  \centerline{$r$ - radius, $r_{in}$ - inner radius, 
              $r_{out}$ - outer radius}
  \centerline{$h$ - height, $w$ - width, $d$ - depth, $l$ - length} 
  \vspace*{-0.30in}
\end{table}
%

%
%
The value of using primitive shapes is in the ability to automatically
synthesize a vast library of shapes through gridded sampling within the
parametric domain of each class.  Table \ref{tab:primtive_shape_design}
lists the parameter coordinates (middle column) for each primitive
shape class. Sections \ref{datagen} and \ref{domain_alignment} detail the
training method used to synthetically generate depth images and known
segmentations from the parametric shape classes. 
Once trained, the Mask-RCNN \cite{he2017mask} network decomposes an input
depth image into a set of segmentations reflecting hypothesized
primitive shapes, as shown in Figure \ref{fig:pipeline}(a).  Segmentations
for different input depth images overlaid on the corresponding, cropped RGB
images are demonstrated in Figure \ref{fig:segdemo}\,. 
The color coding
is 
red: {\em cylinder}, 
orange: {\em ring}, 
green: {\em cuboid},
yellow: {\em stick}, 
purple: {\em semi-sphere}, 
and
blue: {\em sphere}.
For individual,
sparsely distributed, and 
clustered objects, 
the segmentation method captures the primitive shape regions of the sensed
objects. 


%
%
%
%
%

\subsection{Shape Extraction and Estimation, and Grasp Family}
\label{shapeTograsp}

%
%
%
Given a target object region to grasp, the intersecting shape primitive
regions are collected and converted into separate partial point clouds.
Each point cloud needs to be associated to a parametric model of the known
shape.  Using an object and grasp database of examplar shapes, a
multi-start Iterative Closet Point (ICP) algorithm matches the exemplar
shapes to the point cloud. The multiple starting points prevent local
minima issues by providing different orientation guesses for the initial
estimate.  The final match with the lowest error
score is selected as the object model for hypothesizing grasps. 
In addition to the model, the transformation aligning the point cloud
with the shape is kept for mapping grasps to the world frame.
This matching step is depicted in Figure \ref{fig:pipeline}(b).  
Given the identified type of shape and its corresponding shape parameters,
one or more families of grasps are recovered. 
The geometric primitives do not need to match the target region exactly. 
As long as the error between the object and the best matching primitive
shapes is small relative to the gripper approach properties.

\begin{figure}[t]
  \centering
  \vspace*{0.06in}
  \begin{tikzpicture}[inner sep=0pt, outer sep=0pt]
    \node[anchor=south west] (CyW) at (0in,0in)
      {\includegraphics[height=0.8in,clip=true,trim=1.5in 0.65in 1.5in 0.50in]{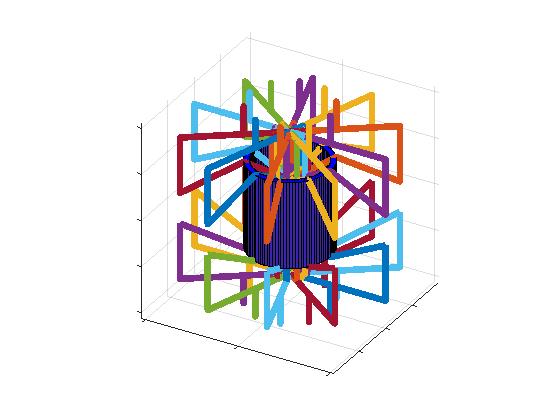}};
    \node[anchor=south west,xshift=5pt] (CyT) at (CyW.south east)
      {\includegraphics[height=0.8in,clip=true,trim=1.5in 0.65in 1.5in 0.50in]{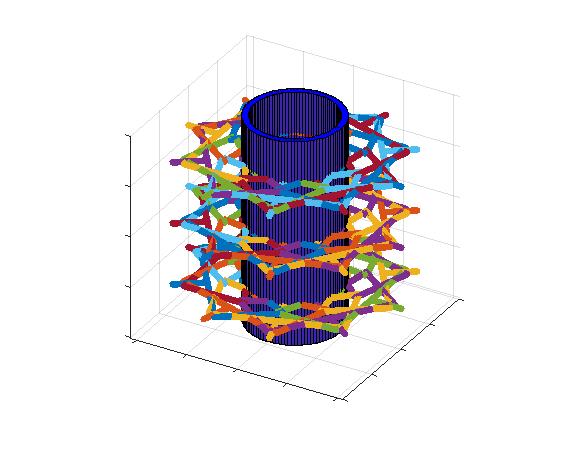}};
    \node[anchor=south west,xshift=5pt] (SS) at (CyT.south east)
      {\includegraphics[height=0.8in,clip=true,trim=1.5in 0.65in 1.5in 0.50in]{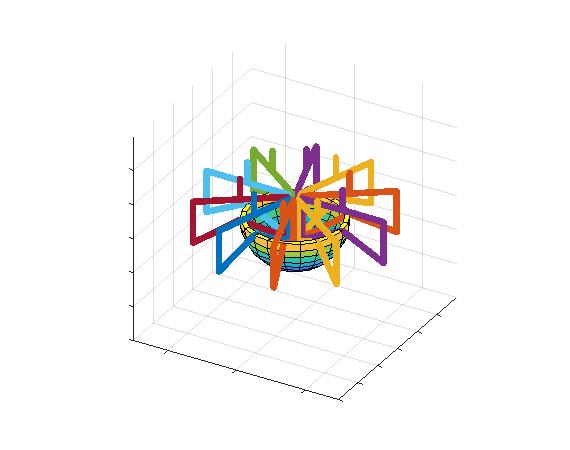}};
  \end{tikzpicture}
  \caption{Grasp family for wide cylinder, tall cylinder, and semi-sphere. 
    \label{fig:grasp_family}}
  \vspace*{-0.20in}
\end{figure}

%
%
Each object class has a family of grasps based on its geometry, with
each member of the family corresponding to a set of grasps associated to
transformation by a group action due to the symmetry of the primitive.
For grabbing wide cylinders, there are two members corresponding to
grabbing from the top or from the bottom, with the free parameter being
rotation about the cylinder axis. 
For grabbing a thin cylinder, there is one two-parameter set corresponding
to translation along or rotation about the cylinder axis.
For the tall cuboid there are four sets in the family, each one associated
to grasping from one edge and translating parallel to the edge.  Example
grasps from these described families are depicted in Figure \ref{fig:grasp_family}.
Under ideal conditions (i.e., no occlusion, no collision, and reachable)
all grasps are possible (e.g., the object is floating). In reality, only a
subset of all possible grasps will be feasible. Using the shape primitive 
grasp family avoids the problem of incomplete annotations or sparse
sampling from regions of the grasp space.
Since the geometry of each shape is known, the predicted grasp family is
robust (modulo the weight distribution of the object). 

\newcommand{\quatR}{\mathfrak{q}}
\newcommand{\score}{s}
\newcommand{\scoreR}{s_{rot}}
\newcommand{\scoreT}{s_{trans}}
\newcommand{\costR}{C_{rot}}
\newcommand{\costT}{C_{trans}}
\newcommand{\rankMC}{\gamma_{MC}}
\newcommand{\rankTS}{\gamma_{TS}}
%
%
%
\subsection{Grasp Prioritization and Selection} \label{ranking_algorithm}
%
%
The final step prior to execution is to select one grasp from the set of
candidate grasps as the one to use. 
DexNet 2.0 \cite{mahler2017dex} was explored as a means to score the grasps,
but performance degraded for the angled camera perspective of the setup here. 
Instead, a simple geometric grasp prioritization scoring function was used
(unrelated to existing grasp quality scoring functions).
It considers the required pose of the hand relative to the world frame,
$\grasp = \grasp^{\mathcal{W}}_{\mathcal{H}}$, which is located at the
manipulator base.
The prioritization scheme prefers the desired grasp to minimize translation
and favor approaching from above. The first collision free grasp, when
tested in the prioritized ordering, is the grasp selected.

%
The grasp prioritization score will consist of contributions from the
translational and rotational components a test grasp pose. The
translation score contribution will depend on the length of the translation
element (i.e, the distance from the world/base frame). Define the
translation cost to be $\costT(T) = \|\|T\|\|$ where $T$ is the translation
interpreted to be a vector in $\mathbb{R}^3$.
The rotational contribution regards the equivalent quaternion
$\quatR \in SO(3)$ as a vector in $\mathbb{R}^4$ and applies the following
positive, scalar binary operation to obtain the orientation grasp cost:
\begin{equation}
  \costR(\quatR) = \vec{w} \odot \vec \quatR, \quad \text{where}\ 
    \vec w = (0, \omega_1, \omega_2, \omega_3)^T\!\!\!,\ \omega_i > 0,
\end{equation}
with
$\vec a \odot \vec b \equiv |a^1 b^1| + |a^2 b^2| + |a^3 b^3| + |a^4 b^4|$.
This cost prioritizes vertical grasps by penalizing grasps that do not
point up/down.
Alternative weightings are possible depending on the given task, or the
robot to workspace configuration.

%
%
%
The costs are computed for all grasp candidates, then converted into a
score by normalizing them over the range of obtained scores,
\begin{equation} \label{rt_score_eqt}
\begin{split}
  \scoreR(\costR) & = 1 -  \frac{\costR - \costR^{min}}
                                {\costR^{max} - \costR^{min}} \\
  \scoreT(\costT) & = 1 -  \frac{\costT - \costT^{min}}
                                {\costT^{max} - \costT^{min}} 
\end{split}
\end{equation}
where the $\cdot^{min}$ and $\cdot^{max}$ superscripts denote the min and
max over all scores grasps.
Two methods are tested for combining $\scoreR$ and $\scoreT$ into a
final grasp prioritization score: 

\begin{hangparas}{1em}{1}
- \textbf{Mixed Criteria Score Ranking (MC):}
    Simply generate the weighted sum of the two scores,
    \begin{equation} \label{mscr}
      \rankMC(\scoreR, \scoreT) = \lambda_{R} \scoreR + \lambda_{T} \scoreT
    \end{equation}
    for $\lambda_R,\lambda_T>0$.

- \textbf{Two Stage Filtered Ranking (TS):}
    The two stage approach first selects the top $n$ grasps based on their
    translation score $\scoreT$, then re-ranks them based on the rotation
    score $\scoreR$. Denote this ranking method by $\rankTS$.
\end{hangparas}

\noindent
After ordering the grasps according to their grasp prioritization score,
the actual grasp applied is the first one to be feasible when a grasp plan
is made from the current end-effector pose to the target grasp pose.
This third and final step in the grasp identification process is shown 
Figure \ref{fig:pipeline}(c).

%
%

\begin{figure}[t]
  \centering
  \vspace*{0.06in}
\tikzstyle{vecArrow} = [semithick, decoration={markings,mark=at position 1
with {\arrow[scale=1.25,semithick]{open triangle 60}}},
   double distance=3.4pt, shorten >= 6.75pt,
   preaction = {decorate},
   postaction = {draw,line width=3.4pt, white,shorten >= 6.5pt}]
\tikzstyle{innerWhite} = [semithick, white,line width=1.4pt, shorten >=
6.25pt]

  \begin{tikzpicture}[inner sep = 0pt, outer sep= 0pt]
    \node[anchor=south west] (SimOrig) at (0in,0in)
      {\includegraphics[width=0.15\textwidth]{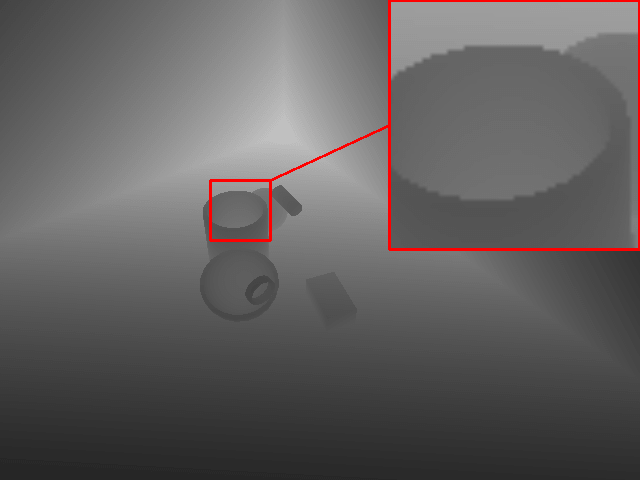}};
    \node[anchor=south west, xshift=0.75in] (SimFilt) at (SimOrig.south east)
      {\includegraphics[width=0.15\textwidth]{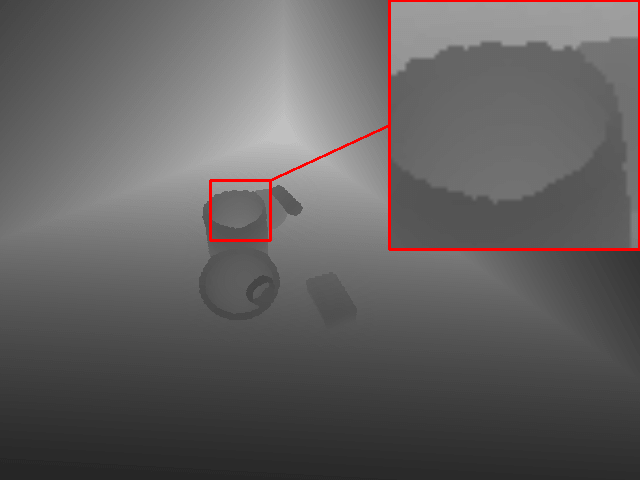}};

    \node[anchor=south west, yshift=0.1in] (KinOrig) at (SimOrig.north west) 
      {\includegraphics[width=0.15\textwidth]{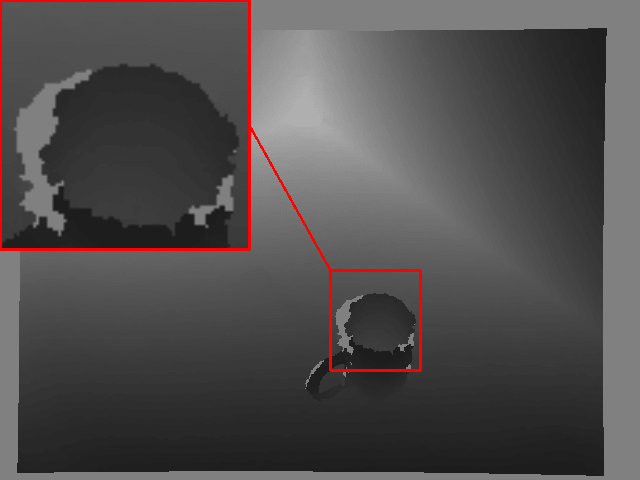}};
    \node[anchor=south west, xshift=0.75in] (KinFilt) at (KinOrig.south east) 
      {\includegraphics[width=0.15\textwidth]{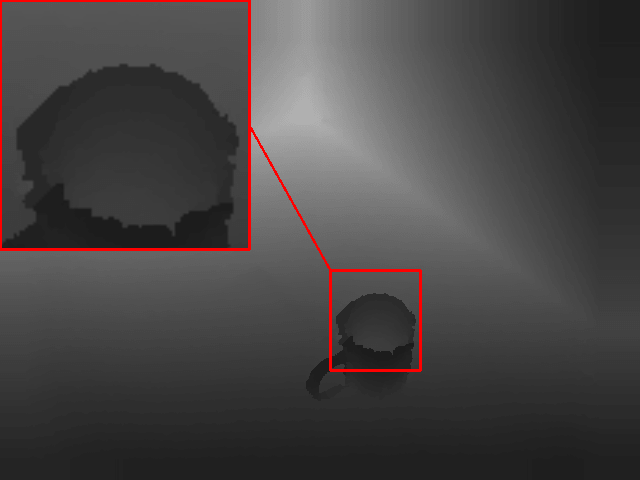}};

    \node[anchor=south west,xshift=2pt,yshift=2pt] at (SimOrig.south west)
      {\footnotesize \textcolor{white}{Simulated}};
    \node[anchor=south west,xshift=2pt,yshift=2pt] at (KinOrig.south west)
      {\footnotesize \textcolor{white}{Kinect}};
    \node[anchor=base,rotate=90,yshift=7pt] (clean) at (SimOrig.west)
      {\small Too Clean};
    \node[anchor=base,rotate=90,yshift=7pt] (noisy) at (KinOrig.west)
      {\small Too Noisy};
    \node[anchor=north,rotate=90,yshift=-7pt] at 
      ($(KinFilt.east)!0.5!(SimFilt.east)$) {\small Better Matched};

    \draw[vecArrow] ([xshift=5pt]SimOrig.east) to 
      node[above,xshift=-2pt,yshift=5pt,anchor=base]{\small Corrupt} 
      ([xshift=-5pt]SimFilt.west);
    \draw[vecArrow] ([xshift=5pt]KinOrig.east) to 
      node[above,xshift=-2pt,yshift=5pt,anchor=base]{\small Denoise} 
      ([xshift=-5pt]KinFilt.west);

  \end{tikzpicture}
  \caption{Bi-directional image filtering to align training data and
    real data. An oil painting filter applied to training imagery simulates
    the noise of the Kinect depth sensor.  Temporal averaging and spatial
    median filtering regularize the Kinect depth image during run-time.
    \label{fig:oilpainting}}
  \vspace*{-0.20in}
\end{figure}
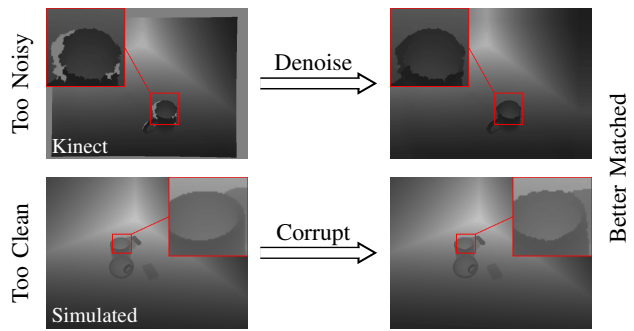

\subsection{Domain Alignment between the simulation and the reality} 
\label{domain_alignment}

State-of-the-art simulators \cite{rohmer2013v,blender} benefit
data generation by automating data collection in virtual environments,
but do so using idealized physics or sensing. Some physical effects are
too burdensome to model. To alleviate this problem, the images from both
sources, the simulation and the depth sensor, are modified to better match.
The objective is to minimize the corrections applied, therefore the first
step was to reduce or eliminate the sources of discrepancies.
Discrepancy reduction involves configuring both environments to match,
which includes the camera's intrinsic and extrinsic parameters, and the
background scene. Comparing images from both sources, the main gap
remaining is the sensing noise introduced by the low-fidelity Kinect v1
depth sensor \cite{planche2017depthsynth,Sweeney2019ASA}.  
The Kinect has occlusion artifacts arising from the baseline between the
active illuminator and the imaging sensor, plus from measurement noise. The
denoising process includes temporal averaging, boundary cropping, and 
median filtering, in that order.
Once the Kinect depth imagery is denoised, the next step is to corrupt the
simulated depth imagery to better match the visual characteristics of the
Kinect. The primary source of uncertainty is at the depth edges or object
boundaries due to the properties of the illuminator/sensor combination. The
simulated imagery should be corrupted at these same locations. The
simulated environment has both a color image and a depth image. The color
image is designed to provide both the shape primitive label and the object
ID, thereby permitting the extraction of object-wise boundaries.  After
establishing the object boundary pixels, they are dilated to obtain an
enlarged object boundaries region, then an oil painting filter
\cite{sparavigna2010cld,mukherjee2014study} corrupts the depth data in
this region.  Considering that manipulation is only possible within a
certain region about the robotic arm, the depth values from both sources
were clipped and scaled to map to a common interval.  Figure
\ref{fig:oilpainting} depicts this bi-directional process showing how it
improves alignment between the two sources.


%
%
\section{Training, Experiments, and Evaluation}

This section describes the automated ground truth synthesis process, the
training process, the robot arm setup, and experimental evaluation
criteria. Code will be open-sourced.

\subsection{Dataset Generation}
\label{datagen}

Based on the hypothesis that a dataset with diverse cominations of primitive 
shapes could induce learning generalizable to household objects, 
the dataset generation procedure consists of the following degrees of
freedom (1) primitive shape parameter; (2) placement order; 
(3) initial $SE(3)$ pose assigned; and (4) mode of placement, for which
there are three modes, {\em free-fall}, {\em straight up from the table-top}, and
{\em floating in the air}.
%
%
%
For simplicity, the shape primitives from Table \ref{tab:primtive_shape_design} 
are reduced to one parameter families after analyzing the household object
databases. Each is denoted by a fixed parameter vector $\rho$ and the
scaling factor $\sigma$, and is described in the third column of 
Table \ref{tab:primtive_shape_design}\,. 
Both cylinders and cuboids are modeled to have a {\em wide-and-short}
category, and a {\em tall-and-thin} category (denoted {\em wide} and {\em
tall} base on the largest parameter) by defining a default parameter set
for each category.  The scaling factor $\sigma$ defines the one parameter
family per class. The {\em ring}, {\em semi-sphere}, and {\em sphere}
vectors are scaled for all coordinates, while the {\em stick} category is
only scaled for the length coordinate $l$. 
When combined with the other domain randomization aspects, the process
samples a sufficiently rich set of visualized shapes once self-occlusion and
object-object occlusion effects are factored in.

For creating instances of the world, the scaling factor is discretized
into 10 steps. Uniform random scalings would have one of the 10 possible
values.  Every image has one instance of each primitive class uniformly
selected from the possible options. The insertion order of the 6 objects
is randomly determined. The initial poses are uniformly randomly
determined within a bounding volume above the table-top. The placement type
is uniformly randomly determined. Through random selection, 100,000 scenes
of different primitive shapes combinations are generated. 
For each instance, RGB and depth images
are collected. Shape color coding provides segmentation ground truth and
primitive shape ID.

\subsection{Data Preprocessing and Training}

Per \S \ref{domain_alignment}, the simulated depth images are re-scaled
then corrupted by a region-specific oil-painting filter.  To align with the
input of Mask R-CNN, the single depth channel is duplicated across the
three input channels.  The ResNet-50-FPN as the backbone is trained from scratch on corrupted dpeth images in PyTroch 1.0
It runs for 100,000 iterations with 4 images per mini-batch. The
primitive shape dataset is divided into a 75\%/25\% training and testing
splits.  The learning rate is set to 0.01 and divided by 10 at iterations
25000, 40000, and 80000. The workstation consists of a single NVIDIA
1080Ti (Pascal architecture) with cudnn-7.5 and cuda-9.0. Dataset
generation takes 72 hours, and training takes 24 hours (4 days total).

\subsection{Robotic Arm Experimental Setup and Parameters}
\label{primitive_shapes_grasping}

\begin{figure}[t]
  \vspace*{0.06in}
  \centering
  \begin{tikzpicture}[inner sep=0pt, outer sep=0pt]
    \node[anchor=south west] (Robo) at (0in,0in)
      {\includegraphics[width=0.3\textwidth]{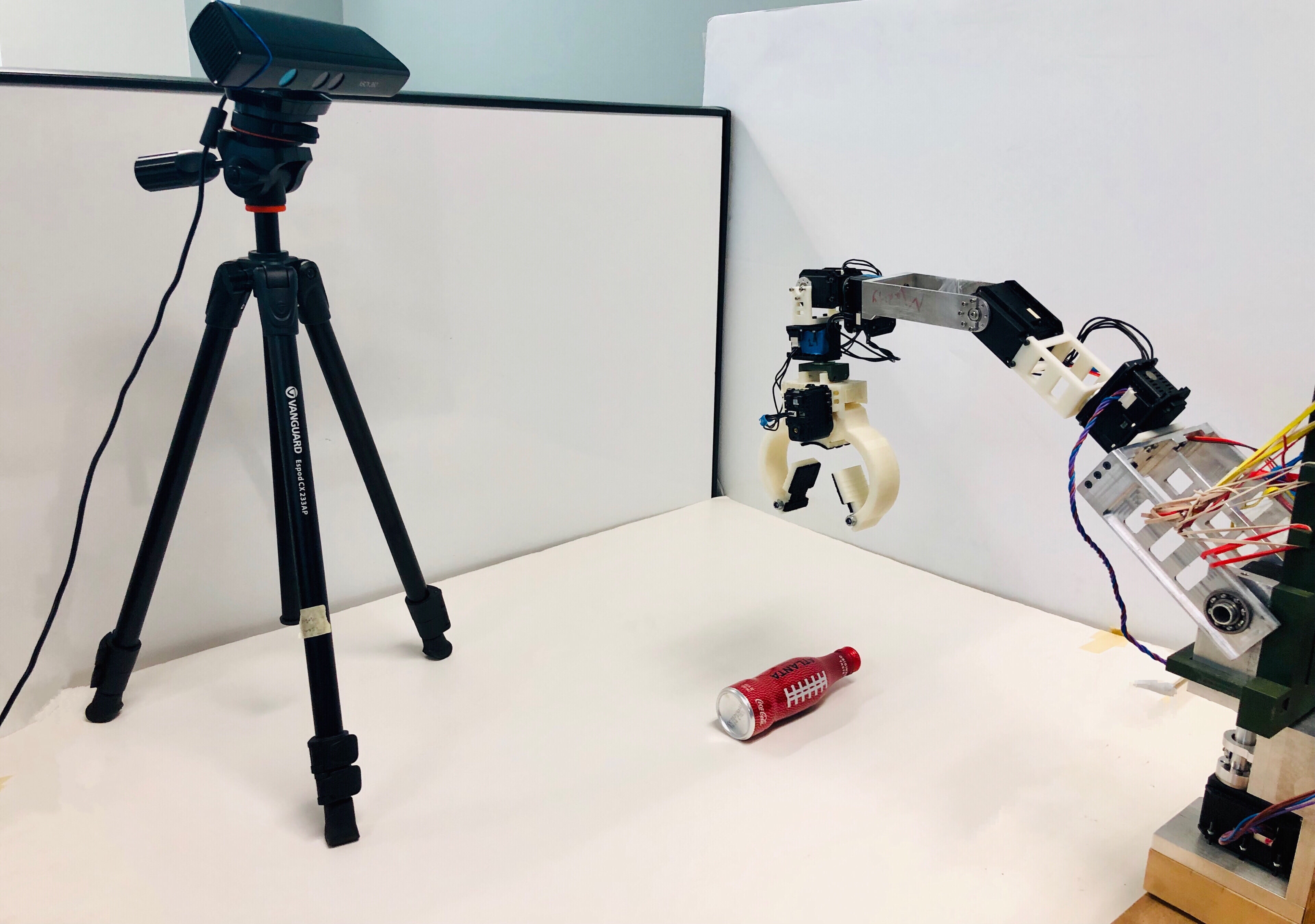}};
    \node[anchor=south west,xshift=3pt] (Actual) at (Robo.south east)
      {\includegraphics[height=0.72in]{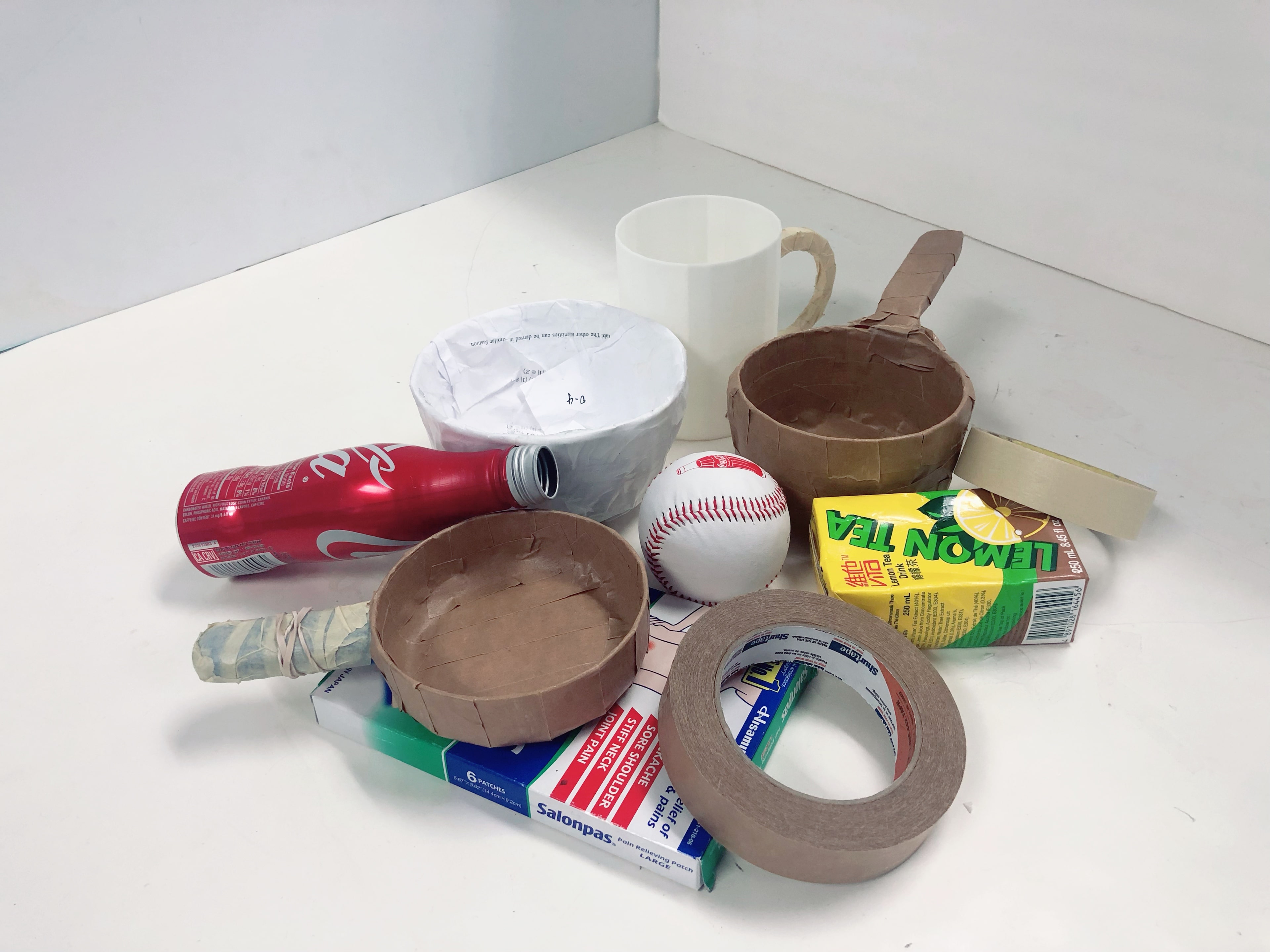}};
    \node[anchor=north west,xshift=3pt] (Printed) at (Robo.north east)
      {\includegraphics[height=0.72in]{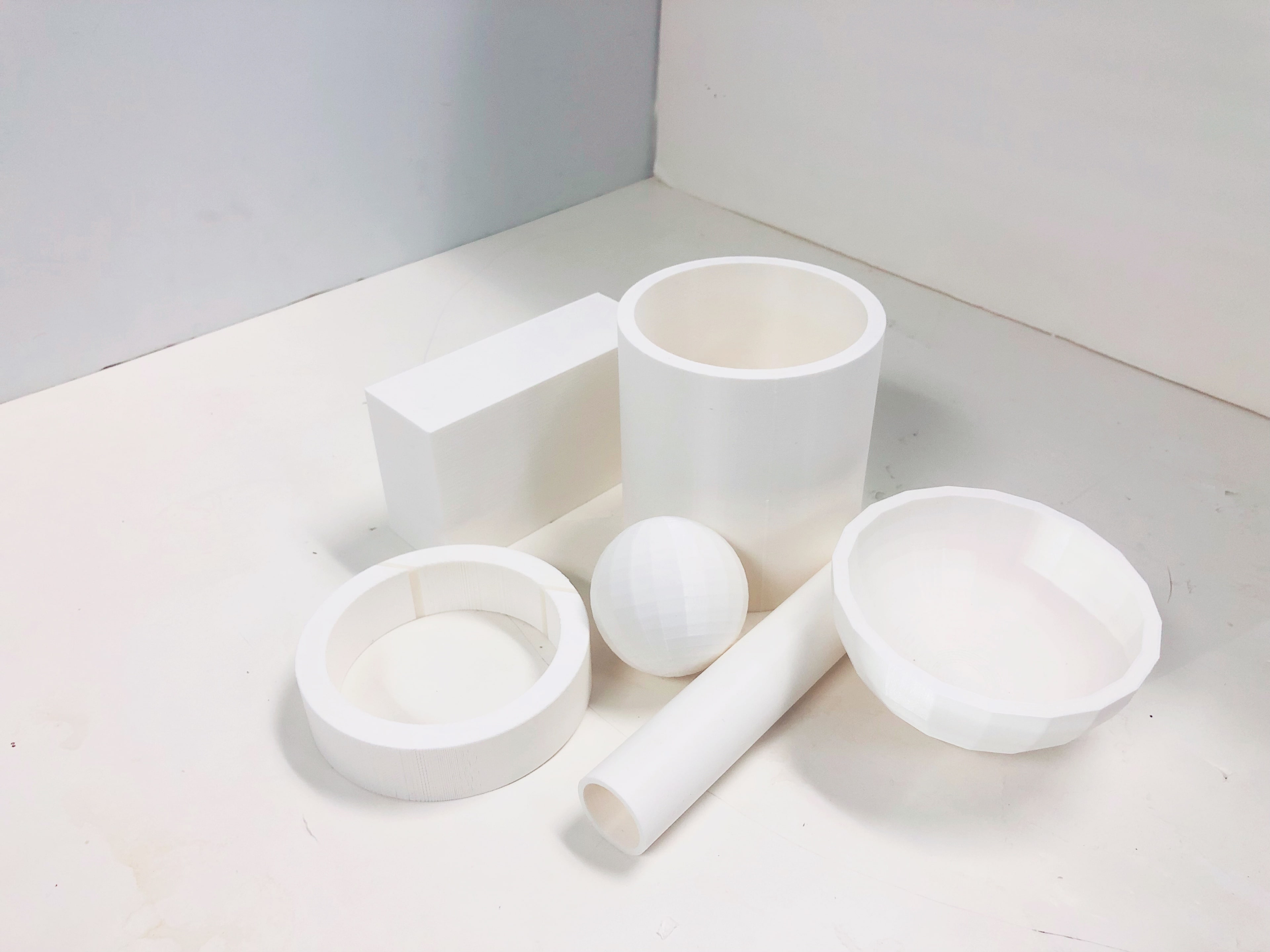}};
  \end{tikzpicture}
  \caption{Experimental setup and images of known and novel object sets.
    \label{fig:exp_setting}}
  \vspace*{-0.20in}
\end{figure}

The robotic arm and RGB-D camera setup used for evaluating the outcomes of
the proposed pipeline is shown in Figure \ref{fig:exp_setting} (left).  It
is eye-to-hand. The camera to manipulator base frame is established based
on an ArUco tag captured by the camera. Both the described method and the
implemented baseline methods are tested on this setup. 
A set of 3D printed shapes designed to fit the training shapes is used for
{\em known objects} testing and evaluation, Figure \ref{fig:exp_setting}
(top right).
An additional set of objects is subsequently used for {\em novel objects}
testing and evaluation, Figure \ref{fig:exp_setting} (bottom right).

The primitive shape and grasp family database is populated by selecting 10
exemplars from each primitive shape type.  Each continuous grasp set is
likewise discretized according to the dimensions of our gripper so that
neighboring grasps are not too similar. 
For MC ranking, the weights in \eqref{mscr} are set equal, 
$\lambda_R = \lambda_T = 0.5$.  For TS ranking, the top 10 are chosen.
Collision checking for grasp feasibility is done by the {\em Planning
Scene} module in {\em MoveIt!} \cite{chitta2012moveit}.  
Open-loop execution is performed with the plan of the top grasp.

%
%
\subsection{Evaluation Metrics}
Evaluation of the proposed pipeline consists of testing on novel input data
as purely a visual recognition problem, followed by testing on the
experimental system. For the visual segmentation evaluation, the
segmentation accuracy is computed by $F_{\beta}^{\omega}$ \cite{Margolin2014HowTE}:
\begin{equation} \label{F_measure}
  F_{\beta}^{\omega} = {(1 + \beta^{2})}
    \,{{Pr^{\omega}\cdot Rc^{\omega}}
    \Big/
    {(\beta^{2}\cdot Pr^{\omega}+Rc^{\omega})}}.
\end{equation}
where $Pr$ is precision and $Rc$ is recall, $\beta = 1$, and $\omega$ is a
Gaussian smoothing factor.
For the robotic arm testing, only the final outcomes of the grasping tests
are scored.  Each experiment consists of 10 repeated runs for a given
configuration. A run or attempt is considered to be a success if the target
object is grasped, lifted, and held for at least 10 seconds. The scoring
metric is the success rate (percentage).

    %

%
%
\section{RESULTS}
%
%

This section details the experiments performed and their outcomes, starting
with a segmentation output test then several manipulation tests.  
The first manipulation test is an in-class grasping test consisting of
inidivudal objects whose shape matches one of the primitive shape classes.
The second is an out-of-class grasping test consisting of novel objects,
some of which composed of multiple shape classes.  Grasping outcomes for
individual objects are quantitatively compared to other methods on the same
test, as well as compared to published methods for similar tests. The third
experiment is a task-oriented grasping task where a specific primitive shape
must be grasped, as would be done when performing a specific task with the
object.  The fourth is a stress test of the system: a multi-object
clearing task.

The baseline implementations are \cite{jain2016grasp} and \cite{Chu_Xu_Vela_2018}.
The method in \cite{jain2016grasp} is a primitive-shape-based system for 
household objects, while \cite{Chu_Xu_Vela_2018} is a publicly available
deep network RGB-D grasping approach. 
The implementation \cite{jain2016grasp} was reproduced, with the
implemented primitives being spherical, cylindrical and box-like primitives. 
Each object is approximated by the best fitting primitive shape.


\subsection{Vision}


Segmentation tests are performed on a set of 3D printed primitive shapes
and compared to manual segmentations. The tested implementations include
a network trained with the original simulated images (no corruption) and
with the oil-painting corrupted images. Per Table \ref{tab:testseg}\,, 
the $F_{\beta}^{\omega}$ segmentation accuracy improved from 0.835 to 0.872
(ranges over $[0,1]$),
with the primary improvement sources being for the
{\em sphere} shape class followed by the {\em stick class}. The
segmentation accuracy is sufficient to capture and label significant portions of 
an object's graspable shape regions, see Figure \ref{fig:segdemo}.

\begin{table}[t]
  \vspace*{0.06in}
  \caption{Performance of PS-CNN on 3D printed primitive shapes\label{tab:testseg}}
  \vspace*{-0.05in}
  \centering
  \setlength\tabcolsep{4pt}
  \begin{tabular}{|l|c|c||c|c|c|}
    \hline
                & Original & Corrupted &
                & Original & Corrupted 
    \\ \hline
    Cylinder    & 0.918    & 0.917 & 
    Ring        & 0.903    & 0.904 
    \\ \hline
    Cuboid      & 0.822    & 0.824 & 
    Stick       & 0.787    & 0.838 \\ \hline
    Semi-sphere & 0.919    & 0.913 & 
    Sphere      & 0.661    & 0.835 \\ \hline
    \multicolumn{3}{|l|}{} &
    {\textbf All}
                & 0.835    & 0.872 
    \\ \hline
  \end{tabular}
%
%
  \caption{Primitive Shapes Grasping (Known) \label{tab:primitive_shape_exp}}
  \vspace*{-0.05in}
  \begin{tabular}{|l|c|c|c|c|}
    \hline
    & \PSMC & \PSTS   & \cite{jain2016grasp} & \cite{Chu_Xu_Vela_2018} \\ 
    \hline\hline
    Cuboid
      & 8     & 9    & 8    & 5    \\ \hline
    Cylinder
      & 10    & 10   & 9    & 8    \\ \hline
    Semi-sphere
      & 9     & 10   & 5    & 7    \\ \hline
    Stick
      & 10    & 8    & 7    & 6    \\ \hline
    Ring
      & 9     & 9    & 6    & 5    \\ \hline
    Sphere
      & 9     & 9    & 8    & 8    \\ \hline
    Success (\%)
      & 91.7  & 91.7 & 71.7 & 65.0 \\ \hline
  \end{tabular}
  \caption{Real Objects Grasping (Unkown) \label{tab:real_objects_exp}}
  \vspace*{-0.05in}
  \centering
  \begin{tabular}{|l|c|c|c|c|}
    \hline
    & \PSMC & \PSTS & \cite{jain2016grasp} & \cite{Chu_Xu_Vela_2018} \\ 
    \hline \hline
    Bowl              
      & 10   & 10   & 6    & 7    \\ \hline
    Mug
      & 10   & 10   & 6    & 8    \\ \hline
    Box
      & 10   & 10   & 9    & 9    \\ \hline
    Baseball          
      & 8    & 7    & 8    & 8    \\ \hline
    Tape              
      & 9    & 8    & 9    & 8    \\ \hline
    Bottle            
      & 9    & 6    & 7    & 8    \\ \hline
    Success (\%)     
      & 93.3 & 85.0 & 75.0 & 80.0 \\ \hline
  \end{tabular}
  \vspace*{-0.15in}
\end{table}

%


%
%
%

\subsection{Physical Grasping of Known and Unknown Objects}
\label{real_objects_grasping}

%
%
%
These tests evaluate the ability of the pipeline to pick up an individual
object in the absence of obstacles or clutter.
The outcomes for the {\em known objects} grasping test are presented in
Table \ref{tab:primitive_shape_exp}\,. The two variants (MS and TS) of the 
primitive shape Mask R-CNN segmentation (PS-CNN) approach outperformed two
baseline methods.  Trained purely on primitive synthetic data, the proposed
approach shows the ability to bridge the gap between simulation and
real-world application. While both grasp prioritization schemes gave the
same success rate, they differed with regards to which objects led to more
failures. The baseline \cite{jain2016grasp} also achieved competitive
success rate on shapes easily approximated by spherical, cylindrical and
box-like primitives, but less so for the other shapes. The baseline
\cite{Chu_Xu_Vela_2018} had the lowest success rate, perhaps an indication
of the limitation of using the image-based $SE(2) \times \mathbb{R}^2$
grasp representation.

%
%
The outcomes for the {\em unknown objects} grasp test are presented in
Table \ref{tab:real_objects_exp}. 
Now, there is a difference between the two grasp prioritization schemes
with MC outperforming TS.  It is explained by the filtering effect of the
prioritization mechanism for TS. In prioritizing and selecting a small set
of grasps based on translation, the resulting grasp subset consists of less
successful grasp orientations. A top scoring one may not map to the most
robust grasp possible for the object in question, suggesting that mixed
prioritization is better over sequential. Ultimately, a true grasp quality
scoring implementation is needed.  The baseline methods also showed
improved performance, but the success rate continued to be lower than the
proposed approach. The larger boost for \cite{Chu_Xu_Vela_2018} over
\cite{jain2016grasp} may be attributed to the deep network model being
trained on a real object dataset \cite{cornellgrasp}.

\begin{table}[t]
  \vspace*{0.060in}
  \caption{Grasping Comparison from Published Works
    \label{tab:real_objects_comparison}}
  \vspace*{-0.05in}
  \centering
  \begin{threeparttable}
    \setlength\tabcolsep{4pt}
    \begin{tabular}{|l|c|c|c|c|}
    \hline
      \multicolumn{1}{|c|}{Approach} & 
      Year & 
      \multicolumn{2}{ c|}{Settings} & 
      Success Rate (\%) \\ 
    \hline
    & & \multicolumn{1}{l|}{Objects} & 
      \multicolumn{1}{l|}{Trials} & 
      \\ 
    \hline
    Lenz $et$ $al.$\cite{lenz2015deep} & 2015 
      & 30  & 100 & 84/89\tnote{*}\rule{0pt}{2.2ex}
      \\ \hline
    Pinto {\em et al.} \cite{pinto2016supersizing} & 2016 
      & 15  & 150 & {66}\rule{0pt}{2ex}
      \\ \hline 
    Watson {\em et al.} \cite{watson2017real} & 2017 
      & 10  & -   & 62
      \\ \hline
    DexNet 2.0  \cite{mahler2017dex}  & 2017 
      & 10  & 50  & 80 
      \\ \hline
    Lu $et$ $al$. \cite{lu2018planning} & 2018 
      & 10  &  -  & 84
      \\ \hline
    DexNet 2.0  \cite{mahler2019learning} & 2019 
      & 50  & 5   & 50 
      \\ \hline
    Satish {\em et al.} \cite{satish2019onpolicy} & 2019 
      & 8 & 80    & 87.5
      \\ \hline \hline
    \PSMC
      & - & 10\tnote{**}\rule{0pt}{2.2ex}\rule{0pt}{2ex} 
          & 100  & 94 
      \\ \hline
  \end{tabular}
  \begin{tablenotes}
    \footnotesize
    \item[*] Success rate of 84\%\,/\,89\% achieved on Baxter\,/\,PR2 robot.
     \item[**] Combines the unique objects from Tables 
        \ref{tab:real_objects_exp} and \ref{tab:task_oriented_grasp}\,.
  \end{tablenotes}
  \end{threeparttable}
  \vspace*{-0.15in}
\end{table}

Table \ref{tab:real_objects_comparison} collects success rate statistics
and testing details for other state-of-the-art approaches employing only a
single gripper (no suction-based end-effector). 
Even though the test objects may differ from ours, the general object
classes are similar.
Rough comparison provides some context for the relative performance of the
proposed segmentation-based system on household objects.
Organized sequentially, the top performing results are the earliest (Lenz {\em
et al.}) and the latest (Satish {\em et al.}) approaches, achieving 
below 90\% over up to 100 trials. The proposed approach has a higher
success rate suggesting that identifying primitive shape
regions for identifying candidate grasps is beneficial.

%
%
\subsection{Task-Oriented Grasping on Real Objects}
The value of segmenting objects is that the distinct shape regions may
correspond to grasp preferences based on the task sought to accomplish.
Almost all baselines examined in this paper focused on task-free grasping
(or simply pick-n-place operations). In this set of tests, each connected
primitive shape region is presumed to correspond to a specific functional
part. A subsequent grasping test is performed to compare task-free grasping
versus trask-oriented grasping, where in the latter a specific region must
the grasped. Each object consists of more than one functional part.
Only \cite{jain2016grasp} was tested to serve as a baseline since
it is shape-based.
Table \ref{tab:task_oriented_grasp} reports the performance of the systems
on the grasping tests. The method \cite{jain2016grasp} did not perform as
well as in prior tests on account of the more complex shape profiles of
some of the objects. Since it applies a single shape model, it cannot
perform task-oriented grasping.
Overall, the proposed approach did well but experienced a 20\% drop in
success rate when limited to a specific object part suggesting that more
research should be placed on precision grasping of specific object parts
in addition to general grasping (again, grasp quality scoring would help).
Nevertheless the task-oriented approach scores comparably to some of the
outcomes of Table \ref{tab:real_objects_comparison}, suggesting that strong
task-oriented performance should be possible soon.
Achieving task-oriented grasping permits follow-up research on realizing more
advanced semantic grasping of objects.

\begin{table}[t]
  \vspace*{0.060in}
  \caption{Task-Oriented and Task-free Grasp \label{tab:task_oriented_grasp}}
  \vspace*{-0.050in}
  \centering
  \begin{tabular}{|l|c|c|c|c|}
    \hline
    \multirow{2}{*}{Target Objects}
    \rule{0pt}{2ex} & \multicolumn{2}{|c|}{Task-Free} &
    \multicolumn{2}{|c|}{Task-Oriented} \\
    \cline{2-5}\rule{0pt}{2ex}
      & \cite{jain2016grasp} & Ours  
      & \cite{jain2016grasp} & Ours  \\ \hline
    \hline
    Mug         & 6/10   & 10/10 & -      & 8/10  \\ \hline
    Pot         & 3/10   & 9/10  & -      & 7/10  \\ \hline
    Pan         & 5/10   & 9/10  & -      & 7/10  \\ \hline
    Basket      & 6/10   & 10/10 & -      & 8/10  \\ \hline
    Handbag     & 9/10   & 10/10 & -      & 8/10\rule{0pt}{2.1ex}  \\ \hline
    \hline
    Success Rate (\%)  
                & 58.0   & 96.0  & 0      & 76.0\rule{0pt}{2ex}\\
    \hline
  \end{tabular}
  \caption{Multi-Object Grasping Comparison \label{tab:multi_object}}
  \vspace*{-0.050in}
  \begin{tabular}{|l|c|c|c|c|c|c|c|}
    \hline
    Method & \#Obj. & \#Sel. & \#Trials & TC & S & OC & C \\ \hline
    \cite{pinto2016supersizing}   
      & 21 & 10 & 5  & None & 38 & 100 & 100 \\ \hline
    \cite{gualtieri2016high} 
      & 10 & 10 & 15 & 3Seq & 84 & 77 & -- \\ \hline
    \cite{mahler2017dex}     
      & 25 & 5  & 20 & 5Seq & 92 & 100 & 100 \\  \hline
    \cite{levine2018learning}    
      & 25 & 25 & 4  & 31G  & 82.1 & 99 & 75 \\ \hline

    \cite{ni2019new}  
      & 16 & 8  & 15 & +2 & 86.1 & 87.5 & -- \\  \hline
    \cite{mahler2019learning}
      & 25 & 5  & 20 & 5Seq & 94 & 100 & 100 \\ \hline \hline
     \PSMC & 10 & 3-5 & 10  & +2  & 50 & 80  & 70\rule{0pt}{2ex}  \\ \hline
  \end{tabular}
  \vspace*{-0.15in}
\end{table}



\subsection{Multi-Object Grasping on Real Objects}
The current study is aimed at establishing grasp candidates for individual
objects rather than clutter removal or bin-picking, however additional
testing was performed to gauge the limits of the implementation.
Performance is not expected to be high, since the current version would be
equivalent to DexNet 1.0 or 2.0 in terms of maturity (vs DexNet 2.1).  
From the set of 10 objects, a smaller set of 3 to 5 objects are randomly
selected and placed on the workspace, see Figure \ref{fig:segdemo}\,. 
Following \cite{ni2019new} the robot has $n+2$ grasp attempts to remove the
objects (\cite{levine2018learning} used $n+5$ for $n=25$). We then
calculatd the grasp success percentage (S), the object clearance
percentage (OC), and the completion percentage (C).
Table \ref{tab:multi_object} compares these statistics to other published
works. There are some differences regarding trial termination criteria
(TC), with 
$k$G signifying up to $k$ grasp attempts, 
$+k$ signifying $n+k$ allowed grasps for $n$ objects, and 
$k$Seq meaning $k$ sequential failures. 
As noted earlier, combining the PS-CNN with a downstream grasp quality
CNN (GQ-CNN) trained to recognize grasps from a richer set of camera
perspectives would improve object grasping in clutter. So would closed-loop
operation to correct the tracking error of the low-cost servomotors (the
arm build cost is under \$3k).

%
%
\section{Conclusion}

This paper leverages recent advances in deep learning to realize a shape
primitive segmentation based approach to grasping. Having shape primitive
knowledge permits grasp recovery from known grasp families. It returns to
one of the classical paradigms for grasping and shows that high performance
grasp candidates can be learnt from simulated visual data without
simulating grasp attempts.  The segmentation-based approach permits
task-oriented object grasping in contrast to the current approaches
emphasizing grasp learning.  Robotic grasping experiments indicate a 94\%
grasp success rate in task-free grasping and 76\% for task-oriented
learning.  Future work aims to improve grasping success by exploring the
use of contemporary grasp networks to score the grasp candidates for
robustness or success likelihood, which is what the current approach lacks.
It should then succeed in clutter removal or bin-picking style problems.


\bibliographystyle{IEEEtran}
\bibliography{main.bib}

\end{document}